\documentclass{article}
\usepackage{amsmath,amssymb,amsthm}  
\usepackage{fontawesome5}
\usepackage{tikz}
\usepackage{longtable}
\usepackage{array}

\newcolumntype{P}[1]{>{\raggedright\arraybackslash}p{#1}}

\usetikzlibrary{positioning,arrows.meta,fit,calc,shadows.blur}
\DeclareUnicodeCharacter{2014}{, } 
\definecolor{nuPurple}{HTML}{4E2A84}
\definecolor{nuLight}{HTML}{EEF0F7}
\definecolor{panel}{HTML}{FAFBFD}
\definecolor{teal}{HTML}{00A7A7}
\definecolor{pink}{HTML}{E94E77}
\definecolor{indigo}{HTML}{3B82F6}
\definecolor{okgreen}{HTML}{5AA469}
\definecolor{amber}{HTML}{F59E0B}
\definecolor{softline}{RGB}{210,214,220}
\usepackage{amsmath}
\usepackage{amssymb}
\usepackage{placeins}
\usepackage{wrapfig}
\usepackage{wrapfig}
\usepackage{enumitem}

\theoremstyle{plain}  

\usepackage{placeins}
\theoremstyle{definition} 

\theoremstyle{remark} 
\usepackage{float}

\usepackage{appendix} 
\usepackage{float}
\usepackage{PRIMEarxiv}
\usepackage{algorithm}
\usepackage{algpseudocode}
\usepackage[utf8]{inputenc} 
\usepackage{bbm}

\usepackage[T1]{fontenc}    
\usepackage{hyperref}       
\usepackage{url}            
\usepackage{booktabs}       
\usepackage{amsfonts}       
\usepackage{nicefrac}       
\usepackage{microtype}      
\usepackage{lipsum}
\usepackage{fancyhdr}       
\usepackage{graphicx}       
\graphicspath{{media/}}    
\usepackage{color,soul}
\usepackage{xcolor}
\usepackage{amsmath}
\usepackage{longtable}
\usepackage{tabularx}
\usepackage{array}
\usepackage{tabularx}
\usepackage{array}
\usepackage{booktabs}
\newcolumntype{Y}{>{\raggedright\arraybackslash}X}
\newcolumntype{L}[1]{>{\raggedright\arraybackslash}p{#1}}

\title{From Redaction to Restoration: Deep Learning for Medical Image Anonymization and Reconstruction}

\author{%
\parbox{\textwidth}{\centering
Adrienne~Kline\textsuperscript{1,2,3,4,*}, 
Abhijit Gaonkar\textsuperscript{5},
Daniel~Pittman\textsuperscript{1,2,3,4}, 
Chris Kuehn\textsuperscript{4}, 
Nils~Forkert\textsuperscript{6}, \\
\textsuperscript{1}Center for Artificial Intelligence, BCVI, Northwestern Medicine, Chicago, IL, USA \\
\textsuperscript{2}Department of Electrical and Computer Engineering, Northwestern University, Chicago, IL, USA \\
\textsuperscript{3}Department of Surgery, Northwestern University, Chicago, IL, USA \\
\textsuperscript{4}Xtasis Inc., Chicago, IL, USA \\
\textsuperscript{5}Medtronic, USA\\
\textsuperscript{6}Departments of Radiology and Clinical Neurosciences, University of Calgary, Calgary, AB, Canada.\\
\textsuperscript{*}Corresponding author: \texttt{adrienne.kline@northwestern.edu}
}%
}

\begin{document}
\maketitle
\begin{abstract}
Removing patient-specific information from medical images is crucial to enable sharing and open science without compromising patient identities. However, many methods currently used for deidentification have negative effects on downstream image analysis tasks because of removal of relevant but non-identifiable information. This work presents an end-to-end deep learning framework for transforming raw clinical image volumes into de-identified, analysis-ready datasets without compromising downstream utility. The methodology developed and tested in this work first detects and redacts regions likely to contain protected health information (PHI), such as burned-in text and metadata, and then uses a generative deep learning model to inpaint the redacted areas with anatomically and imaging plausible content. The proposed pipeline leverages a lightweight hybrid architecture, combining CRNN-based redaction with a latent-diffusion inpainting restoration module (Stable Diffusion 2). We evaluate the approach using both privacy-oriented metrics, which quantify residual PHI and success of redaction, and image-quality and task-based metrics, which assess the fidelity of restored volumes for representative deep learning applications. Our results suggest that the proposed method yields de-identified medical images that are visually coherent, maintaining fidelity for downstream models, while substantially reducing the risk of patient re-identification. By automating anonymization and image reconstruction within a single workflow, and dissemination of large-scale medical imaging collections, thereby lowering a key barrier to data sharing and multi-institutional collaboration in medical imaging AI.
\end{abstract}


\vspace{-4mm}
\section{Introduction}
Machine learning models trained using this medical imaging data have considerably contributed to the speed and repeatability of diagnostic tasks, sometimes even outperforming human experts in assessments of chest x-rays, retinal scans, and echocardiography assessments \cite{patel2019human} \cite{kamran2019optic} \cite{madani2018fast}. However, training machine learning models on small or noisy datasets calls limits the reproducibility and generalizability of these models for real-world deployments. Large conglomerate imaging datasets have facilitated open-source repositories of information for the development of generalizable machine learning and artificial intelligence models for computer-aided diagnosis support and clincial decision making \cite{ADNI2022}\cite{OAI2022}\cite{johnson2024mimiccxr}. In fact, these databases provide current state-of-the-art benchmarks for the development and testing of novel methodologies \cite{biswas2019state}\cite{hering2022learn2reg}. Therefore, data cleaning and subsequently sharing are a necessary first step for facilitating large imaging repositories which can impact downstream health inferences drawn from algorithmic decisions. Cleaning and preparing medical imaging data often includes outlier detection, removal of incomplete data, imputation, and deidentification (metadata and pixel-level PHI). The latter is the focus of the current study. Imaging data sets contain both relevant and irrelevant information (e.g., redundant, private).

\begin{figure}
    \centering
    \includegraphics[width=0.35\linewidth]{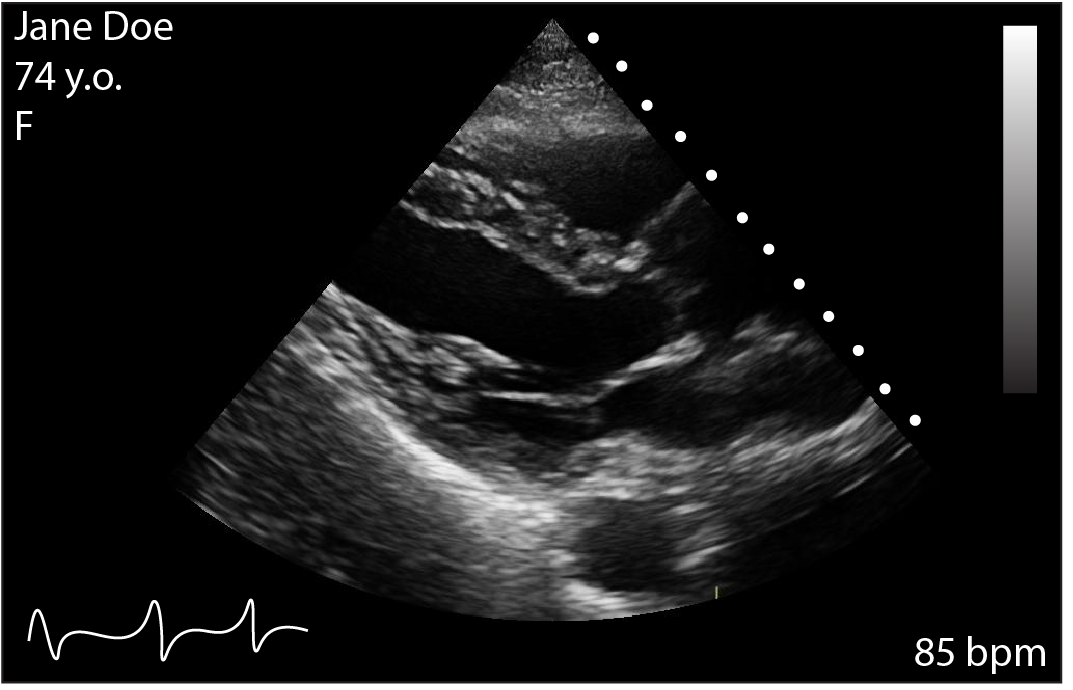}
    \caption{Typical example of an echocardiogram with a synthetic PHI overlay}
    \label{fig:enter-label}
\end{figure}

The aims of the current work are threefold: 1) Provide a framework for detecting and extracting PHI-based text from medical images and increasing security; 2) Apply and compare various methods of inpainting for redacted areas of images to meaningfully reconstruct lost pixel information and 3) Compare the influence the reconstruction techniques have on downstream supervised classification tasks. 

\subsection{State-of-the-Art}
Due to the risk of unauthorized sharing of protected health information (PHI), efforts have been made to de-identify medical image pixel and metadata for use in broader applications. For example, Newhauser et al. proposed to automate DICOM de-identification using Optical Character Recognition (OCR) to detect and mask all text within the image arrays \cite{newhauser2014anonymization}. Another previously approach to text masking uses a Convolutional Recurrent Neural Network (CRNN). This architecture is a combination of a Convolutional Neural Network (CNN), Recurrent Neural Network (RNN), and Connectionist Temporal Classification (CTC) and is used for image-based sequence recognition tasks for text detection \cite{shi2016end}. More generally, attributes within the image such as curves, lines, intersections, and loops, and how they occur in sequence relative to one another at different scales are often used for this purpose. For example, Fezai et al. applied a convolution neural network architecture using a simple auto-encoder to classify magnetic resonance imaging (MRI) equipment \cite{fezai2023deep} from the metadata contained in the image header. Monteiro et al. proposed a convolution neural network (CNN) pipeline for de-identification of ultrasound DICOM images that removes identifying text only, while leaving non-identifying text intact \cite{monteiro2017identification}. Their software is available as a software-as-a-service, with an anonymization success rate of 89.2\%. Huang et al. presented a method following HIPAA privacy rules, which also implemented an OCR \cite{huang2009privacy} with a recorded success rate of 65\%. Their algorithm, however, struggled to distinguish characters such as "Bg", from "B9" or "B1 from "Bl" and "B0 from "Bo". This may be important depending on downstream NLP-based filtering tasks. Oftentimes this is a difficult task due to nuisance features arising from the noisy medical image background, or text that ovarlies anatomical structures, which complicates the OCR methodology. However, completely de-identifying images is not always practical or needed, depending on the secondary application \cite{parker2021canadian}. Therefore, the aim of de-identification should be to retain meaningful metadata (\textit{e.g.,} age, race, sex in separate files in addition to crosswalks) and pixel (anatomical) information to allow for translation applications, while reducing PHI-induced security risks.

Building on the methodologies and limitations of previous work, we also make use of OCR for text detection with respect to pixel information. However, unlike previous work, we hypothesize that removing and sequestering all text to a separate file structure provides a two-fold advantage. First, by removing all text, we increase assurance that PHI is, in fact, removed from the image content. Second, it places relevant patient information in a format that is easily accessible to end users. Second, extracting ’burned-in’ redundant information (as it is repeated in every frame) from the image array permits the truncation of pertinent image information. For example, oftentimes in ultrasound images text is repeated across frames. This process also decreases the amount of redundant information stored within the pixels, which is an expensive data format, exchanging it for data stored in tabular format. Further, downstream deep learning models may exploit any available text as a shortcut, creating spurious correlations.

Pixel-level redaction of PHI is an essential requirement for protecting privacy in images, particularly in medical imaging. However, naïvely masking pixels (for example, by black boxes or heavy blurring) can affect local intensity distributions, textures, and anatomical boundaries in ways that degrade the performance of downstream computer vision models via shortcuts and biases. However, blurred images may be invertible using super-resolution-like methods. Generally, previous research on visual privacy preservation has shown that there is a fundamental privacy–utility trade-off: stronger obfuscation often comes at the cost of reduced recognition or segmentation accuracy unless the transformation is explicitly designed to preserve task-relevant image content \cite{ravi2024review}. In radiation oncology, for instance, systematic evaluation of defacing algorithms for head-and-neck CT has shown measurable reductions in deep-learning organ segmentation performance for structures near the redacted regions, underscoring that de-identification operations can alter clinically important features if not carefully controlled \cite{ammar2025evaluating}. To mitigate these effects, recent research has turned to deep-learning–based and conditional inpainting methods that aim to reconstruct anatomically plausible content in masked image regions, restoring realistic tissue appearance and intensity distributions while hiding protected health information \cite{santos2025role}. Systematic reviews of medical image inpainting argue that such structure-preserving reconstruction can maintain or even improve downstream model performance compared with simple masking approaches. Thus, it is essential to treat reconstruction of relevant imaging content as a first-class component of any redaction pipeline, and to validate its impact using the same downstream tasks for which the data will ultimately be used \cite{santos2025role}.

In parallel, transformer models have become the de-facto standard for de-identifying clinical free text. BERT-, RoBERTa-, and domain-specific transformers fine-tuned on i2b2/UTHealth and similar corpora routinely achieve PHI token-level F1 scores $\geq$0.95, substantially outperforming earlier conditional random field (CRF) and Long sort term memory (LSTM) architectures and generalizing to multiple languages and note types \cite{meaney2022comparative}. Radiology reports, pathology reports, and longitudinal EHR summaries can now be de-identified in a largely automated fashion.

These advances  for de-identifying clinical free text are beginning to translate to imaging itself. Recent work from industry and academia couples high-recall object detectors with large multimodal transformers to detect and classify burned-in PHI in DICOM images. For example, Truong et al. proposed a three-stage pipeline, which consists of YOLOv11 for imprint localization, OCR, and GPT-4o-based text analysis, that reaches case-level PHI recall of $\approx$0.999 and instance-level recall of $\approx$0.96 on RadPHI-test and MIDI, while also benchmarking latency and cost trade-offs across GPT-4o, Gemini 2.5, and Qwen-VL. John Snow Labs’ Visual NLP similarly uses transformer-based OCR and visual models to de-identify DICOM and whole-slide (SVS) images at scale, handling both metadata and burned-in overlays within a distributed Spark pipeline \cite{melnyk2023dicomdeidvisualnlp}.

Unlike prior work that either redacts DICOM headers that simply masks burned-in text regions, the pipeline proposed in this work is explicitly designed to detect overlaid text and then restore the underlying anatomy using inpainting, while allowing for redaction or swapping of metadata DICOM fields. Thus, the resulting images approximate the appearance of the true source image data while still being de-identified. By pairing robust OCR/detector stages with a lightweight CRNN/CTC redaction architecture and a diffusion-based inpainting module, we not only localize PHI-bearing regions but also reconstruct intensity patterns and structural continuity, preserving relevant imaging information for downstream tasks, such as segmentation, registration, and classification in a way that tag-only or mask-only approaches cannot. At the same time, our results highlight an important limitation of more aggressive anatomy-removal strategies: removing facial structures, skull, or other “identifying” anatomy may be acceptable for some research settings, but it systematically destroys clinically meaningful context (e.g., extracranial disease, device position, musculoskeletal findings) and alters the statistical structure of the image. For downstream AI/ML models, this effectively short-circuits learning by encouraging networks to fit to a distorted, de-identified image distribution that will never be seen in the clinical routine, undermining generalizability, possibly biasing feature learning toward de-identification artifacts, and potentially degrading performance when models are deployed on real, non-redacted pixel data. The restoration-aware pipeline proposed in this work instead seeks to separate “what must be removed for privacy” (text/PHI) from “what must be preserved for science and care” (native anatomy and image statistics), which is precisely the gap not addressed by current de-identification tools. An overview of these historical methodologies/approaches in consort with their data types and limitations are provided in Table \ref{tab:background_summary}.

\renewcommand{\arraystretch}{1.2}
\small

\begin{longtable}{P{3.6cm}|P{2.5cm}|P{3.1cm}|P{3.9cm}|P{1.7cm}}
\caption{Overview of medical imaging privacy-preserving techniques, data types, evaluation metrics, and limitations}
\label{tab:background_summary}\\
\hline
\textbf{Approach} & \textbf{Data type} & \textbf{Metrics / outcomes} & \textbf{Limitations} & \textbf{Ref.} \\
\hline\hline
\endfirsthead

\multicolumn{5}{l}{\small\textit{Table \thetable\ continued from previous page}}\\
\hline
\textbf{Approach} & \textbf{Data type} & \textbf{Metrics / outcomes} & \textbf{Limitations} & \textbf{Ref.} \\
\hline\hline
\endhead

\hline
\multicolumn{5}{r}{\small\textit{(continued on next page)}}\\
\endfoot

\hline
\endlastfoot

End-to-end de-identification pipeline combining OCR, deep-learning based text detection, and skull-stripping and defacing for multi-modal images and raw k-space
& DICOM MRI and CT, whole-slide images, vendor-specific raw data (e.g., Siemens TWIX)
& PHI text-removal accuracy, proportion of sensitive DICOM tags cleared, defacing / skull-stripping quality, per-study runtime and throughput compared with existing toolchains
& Does not recover anatomy within redacted regions; evaluation emphasizes tag/text clearance and skull-stripping quality more than downstream model performance or fidelity of restored content.
& \cite{rempe2025deid} \\
\hline

A two-stage de-identification process for privacy-preserving medical image analysis. Stage 1: PACS export and systematic DICOM header stripping; Stage 2: attribute-level cleansing with rule-based verification and QA
& CT images in DICOM format
& Completeness of PHI removal (manual review of tags and headers), preservation of essential clinical/technical metadata, PACS and DICOM viewer compatibility
& Primarily metadata- and header-focused; does not address burned-in text or image overlays; no mechanism to restore obscured anatomy; relies on rule-based configuration and manual verification.
& \cite{regan2022twostage} \\
\hline

Project-level GDPR-compliant pipelines across AI4HI projects; standardized DICOM tag profiles, ID pseudonymization, tag removal, and region masking
& Multicentre cancer-related imaging datasets plus linked clinical data
& Number and type of DICOM tags modified, overlap of tag profiles across projects, qualitative assessment of GDPR compliance and operational lessons learned
& Provides guidance of GDPR-compliant workflows rather than a concrete pixel-level de-identification algorithm; largely qualitative evaluation; limited treatment of burned-in text; no reconstruction of masked regions.
& \cite{kondylakis2024ai4hi} \\
\hline

Federated learning frameworks with privacy-preserving mechanisms (e.g., DP, secure aggregation) and uncertainty estimation (Bayesian \& ensemble methods)
& Multi-site medical imaging tasks (classification, detection, segmentation) surveyed across modalities
& Model performance (AUC, Dice, accuracy) under FL vs centralized training, empirical/privacy-leakage analyses, calibration and uncertainty-quality metrics reported in reviewed studies
& Addresses privacy at the model-training level rather than at the image or DICOM-object level; raw images and burned-in text remain unchanged; privacy guarantees are largely theoretical/empirical for gradients, not for released image volumes.
& \cite{koutsoubis2024fl} \\
\hline

Conditional GANs to synthesize realistic brain MRIs with tumors for both data augmentation and anonymization
& Multi-parametric brain MRI (e.g., BRATS, ADNI)
& Tumor segmentation Dice score when training on real vs.\ synthetic vs.\ mixed datasets; visual realism and utility of synthetic images as an anonymized surrogate for real patient data
& Synthetic images may still leak identity trends inherent in the training distribution; not designed as a general-purpose DICOM de-identification tool (no tag handling, no explicit text detection); reconstruction of already redacted regions not considered.
& \cite{shin2018gansynth} \\
\hline

CNN-based verification and retrieval models for re-identifying patients from chest radiographs
& Chest X-ray datasets (ChestX-ray14, CheXpert, COVID-19 CXR collections)
& Verification AUC and accuracy for same-/different-patient classification, retrieval performance (mAP@R, precision@1), cross-dataset generalization of re-identification risk
& Demonstrates re-identification risk rather than offering a de-identification solution; limited to chest radiographs; highlights that naive de-identification (e.g., header stripping alone) is insufficient but does not propose reconstruction or robust mitigation strategies.
& \cite{packhaeuser2022reid} \\
\hline

Cloud-based OCR (Amazon Rekognition) plus medical-NLP PHI detection (Amazon Comprehend Medical) followed by automated redaction of text regions
& Chest X-ray and other radiology images in DICOM, PNG, and JPG formats
& Entity-level confidence scores for detected PHI, configurable redaction thresholds, qualitative visual assessment of correct masking of overlaid identifiers in large image batches
& Proprietary cloud services and external data transfer may be problematic for regulated settings; operates purely at the text/overlay level without modeling anatomical content; does not attempt to reconstruct redacted regions or quantify impact on downstream tasks.
& \cite{wiggins2019awsdeid} \\
\hline

Federated CNN training with differentially private stochastic gradient descent (DP-SGD) applied to decentralized histopathology image analysis
& Histopathology tiles from multi-centre cohorts
& Classification performance (AUC, accuracy) vs non-private baselines at different privacy budgets ($\varepsilon$), robustness of FL under DP noise, communication and convergence characteristics
& Protects model updates with DP but leaves visual PHI and burned-in text untouched; performance degrades at stricter privacy budgets; focused on specific histopathology tasks and does not provide an end-to-end image de-identification and restoration pipeline.
& \cite{adnan2022fldp} \\
\hline

Distributed Visual-NLP pipeline for DICOM and whole-slide images combining transformer-based OCR and visual document models with configurable rule-based redaction of detected PHI in metadata, burned-in pixel data, overlays, and encapsulated documents
& DICOM radiology images (CT, MRI, US, etc.), SVS whole-slide pathology images, and embedded documents (PDF/HTML/text) processed on Spark clusters
& Regulatory-grade OCR and PHI detection performance reported on internal and public DICOM+SVS datasets.
& Proprietary, black-box implementation; peer-reviewed quantitative evaluations are limited; image-quality impact of redaction / inpainting not systematically studied; transformer architectures primarily used for OCR and document understanding rather than anatomy-aware restoration.
& \cite{melnyk2023dicomdeidvisualnlp} \\

\end{longtable}

\section{Material and Methods}
\subsection{Creation of Synthetic Pixel-level PHI}
\begin{figure}
    \centering
    \includegraphics[width=0.5\linewidth]{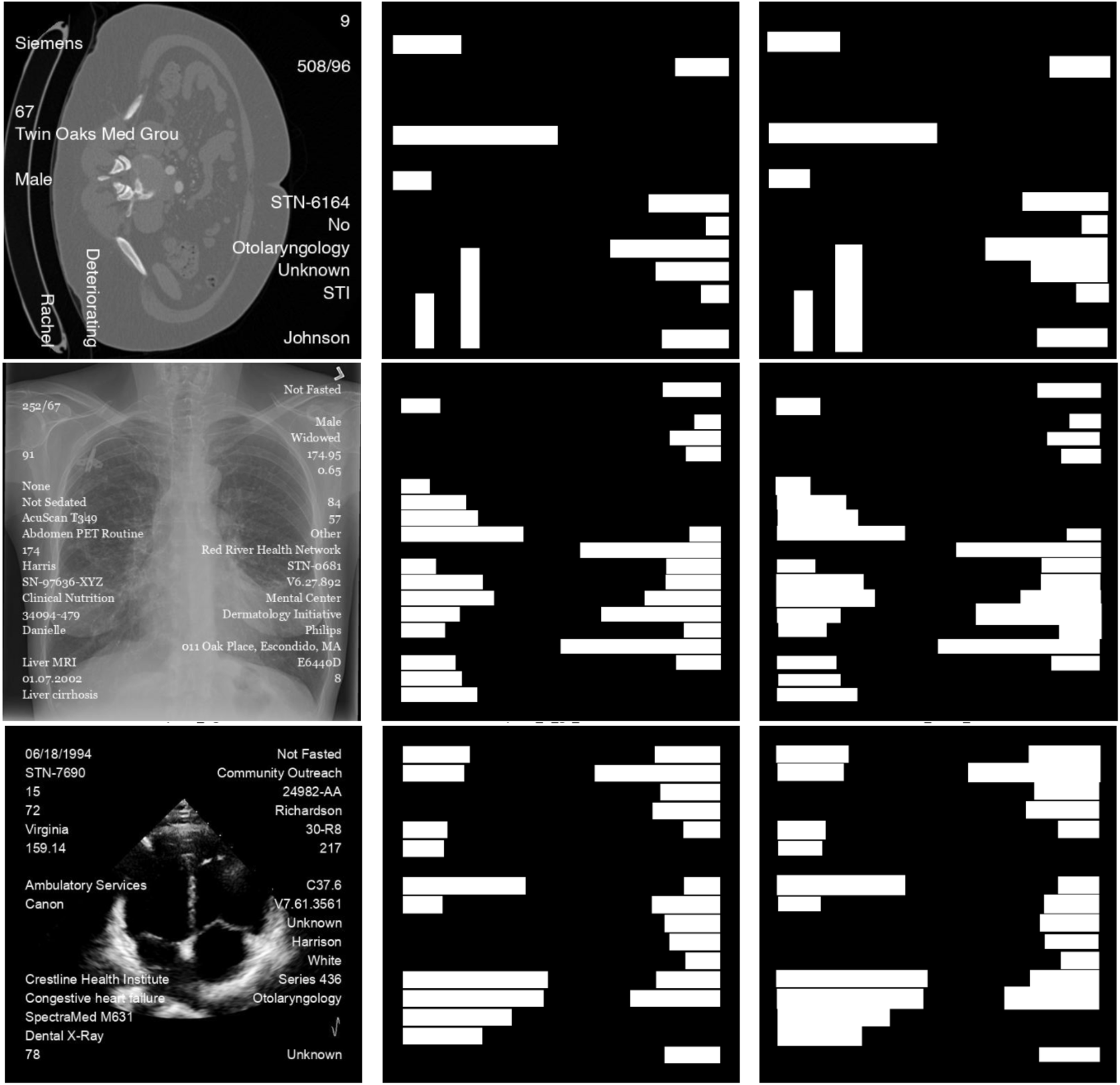}
    \caption{Example data entry of synthetic PHI overlay (left) with ground truth redaction mask (middle) and predicted mask (right).}
    \label{fig: phi detection mask}
\end{figure}

A large and anatomically varied synthetic 2D medical imaging dataset was created (n=29,876) consisting of images from 3D and 2D scans - computed tomography, magnetic resonance, ultrasound, x-ray and angiogram. Therefore, text resembling typical PHI was superimposed on raw images as overlays. Mock text, which ensured all identifiers were fictitious, was fabricated by combining common names, plausible addresses, patterned alphanumeric strings, standard sequence labels, and faux institutional names. Font sizes, spacing, orientation, and text placement were randomized within ranges typical of clinical imaging, ensuring substantial variability of plausible content and composition see Fig \ref{fig: phi detection mask}.

\subsection{Meta-Data Scrubbing}
Metadata was handled using a custom DICOM de-identification pipeline tailored to the established list of 18 HIPAA identifiers \cite{HHS_OCR_HIPAA_Deid_2025}. First, using a list of both historical and active DICOM tags, we manually curated a lookup table linking each DICOM tag (\textit{e.g.,} names, medical record numbers, all elements containing dates or times, geographic locations, and provider identifiers) to whether it constituted a private (vendor-created) DICOM attribute, resulting in several hundred 'HIPAA-compliant' fields. We then applied a metadata redaction/swapping script over all studies that enforced tag-specific actions. Direct identifiers (\textit{e.g., }\texttt{PatientName}, \texttt{PatientID}, \texttt{AccessionNumber}, \texttt{Comments}) were removed or replaced with study-specific pseudonyms; all absolute date and time fields were shifted by a consistent, patient-specific jitter offset to preserve within-patient temporal relationships while obscuring real calendar time; and institution, station, and device/location descriptors were generalized to coarse categories or set to null. For descriptors stored in private or implementation-specific DICOM tags, we used pattern matching on tag names and value content to identify and scrub residual PHI. The resulting headers were revalidated in a second pass by automatically checking that all tags mapped to HIPAA identifiers were either empty, pseudonymized, or date-shifted as intended. A randomly selected subsample (n=500) from the synthetic dataset of (n=29,876) underwent a manual to confirm that no patient, provider, or time/location PHI remained in the metadata.

\subsection{Pixel-level Deidentification}
\begin{figure}
    \centering
    \includegraphics[width=0.55\linewidth]{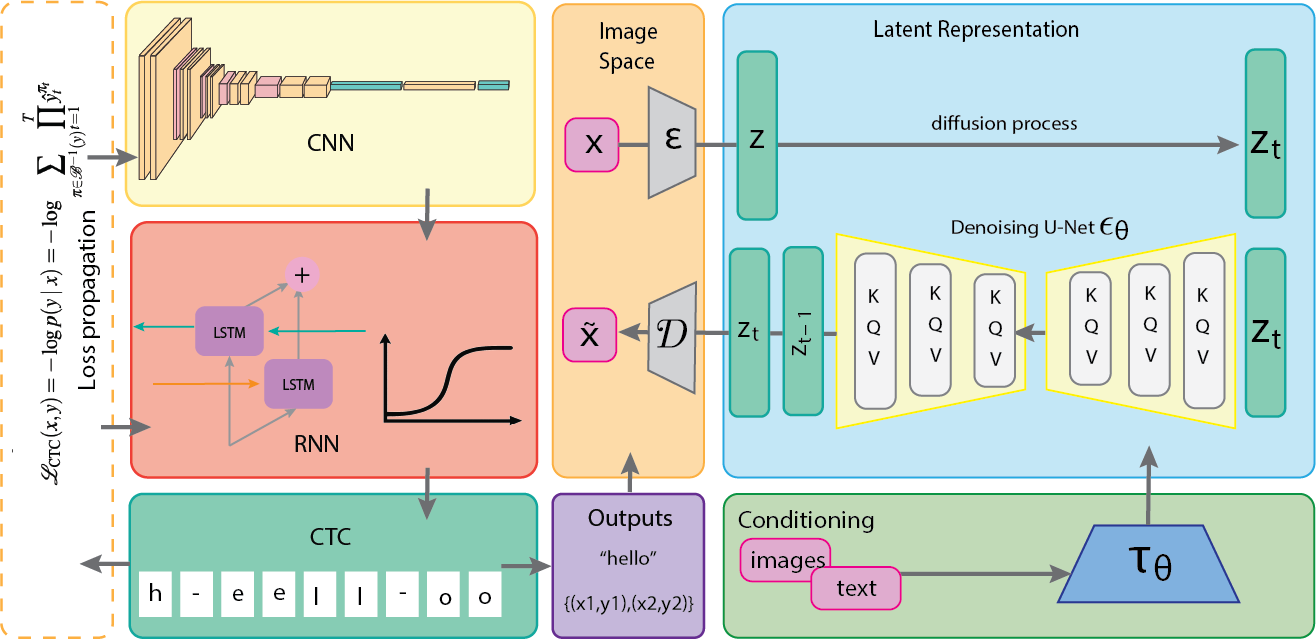}
    \caption{Pipeline and architecture of detecting private health information and subsequent inpainting. The redaction portion of the architecture is composed of a CRNN and CTC, which create a binary that which is directed to a latent-diffusion model for image restoration.}
    \label{fig:architecture}
\end{figure}

An overview of the workflow of the proposed method for pixel-level deidentification and reconstruction is presented in Figure \ref{fig:architecture}. The first step in the pipeline is to detect, extract, and mask text-based data contained within the image arrays. The text recognition architecture is based on a trained convolutional recurrent neural network (CRNN) proposed by Shi et al. \cite{shi2016end} and was trained on various open-source datasets: IC13 \cite{karatzas2013icdar}, IIIT5k \cite{mishra2012scene} and SVT \cite{wang2011end}. IC13 contains 1,015 ground truths of cropped word images, IIIT5k contains 3,000 cropped word test images collected from the internet, and the SVT test dataset consists of 249 street view images collected from Google Street View.

Our PHI detection architecture consists of three main components: a) feature extraction, which is performed using a Visual Geometry Group (VGG19) backbone, b) sequence labeling, which uses an LSTM, and c) decoding, which is performed by Connectionist Temporal Classification (CTC) \cite{graves2006connectionist}.

The convolutional layers in VGG use a small receptive field (3$\times$3, the smallest possible size that still captures left/right and up/down), providing expressiveness, efficiency and depth. There are also 1$\times$1 convolution filters, which act as a linear transformation of the input, which is followed by a ReLU unit. The convolution stride is fixed to 1 pixel so that the spatial resolution is preserved after the convolution. The VGG has three fully-connected layers: the first two have 4096 channels each, and the third has 1000 channels, 1 for each class. All of  VGG’s hidden layers implement a ReLU activation function.

A deep bidirectional Recurrent Neural Network (RNN) is built on top of the VGG backbone to turn images into a sequence of features, and model dependencies across that sequence. The recurrent layers predict a label distribution $y_t$ for each frame $x_t$ in the feature sequence $x = x_1, ..., x_T$. There are several advantages of leveraging recurrent layers. RNNs are well-suited for capturing contextual dependencies within a sequence. In our setting, the “sequence” is the ordered set of features (or symbols) extracted from a single image. Leveraging this within-image context makes image-based sequence recognition more robust than classifying each symbol independently. For example, in text detection, certain characters may require several successive kernels to fully encapsulate the word. A recurrent unit, such as long short-term memory LSTM, is specifically designed to address this problem. Thus, the ambiguity that would otherwise exist in individual characters/kernels is more easily distinguished when observing their contexts — \textit{i.e.,} it is easier to differentiate "n" from "m" by contrasting the character widths than by recognizing each of them separately, in the event downstream NLP-based filtering relies on it. 
Furthermore, an RNN can back-propagate error differentials back to its input (VGG), allowing joint training of the recurrent and the convolutional layers in a unified network.

Lastly, RNNs are designed to process sequences of arbitrary (variable) length by updating a hidden state step-by-step. However, a vanilla RNN often struggles to preserve information over long time spans because gradients can vanish or explode during training. However, an LSTM is a specific type of RNN that addresses this by adding a memory cell and a set of gates that regulate information flow. In a standard LSTM, three multiplicative gates. I.e., input, forget, and output—control what information is written to the cell state, what is retained or discarded, and what is exposed to the hidden state at each step. The memory cell provides a more stable pathway for carrying relevant context forward through the sequence, helping the model learn longer-range dependencies than a basic RNN.

The CTC enables transcription, permitting the training of text recognition models using image and ground truth text pairs. The neural network output matrix encodes text by creating paths that consist of one character per time step. For example, 'ab' or 'aa' represent conceivable paths \cite{liao2022real}. In CTC-based OCR, the network converts the input image into a 1D sequence (often by scanning across the image width) and outputs a $T\times (|V|+1)$ matrix, where each row is a time-step (a position in that sequence) and each column is a character class plus a special blank. The CTC decoder then uses dynamic programming to collapse repeated predictions and remove blanks, producing the final text without requiring pre-segmented characters. This mapping is performed using a dynamic programming approach, which considers all possible alignments between the output probabilities and the target text. During training, the CTC loss function is used to optimize the parameters of the neural network. The CTC loss function calculates the negative log-likelihood of the correct alignment between the output probabilities and the target text, accounting for all possible alignments. Practically, the CTC method allows training of OCR systems using pairs of images and ground truth texts. The ground truth texts are used to calculate the CTC loss function during training, which allows the neural network to learn to output the correct text given an input image.

In this work, we leverage the conditional probabilities using a CTC layer proposed by Graves et al. \cite{graves2006connectionist}. The probability is defined for label sequence $l$ conditioned on the per-frame predictions $y = y_1,..,y_T$, and it ignores the position where each label in $l$ is located. Consequently, using the negative log-likelihood of this probability as the objective function to train the network requires only images and their corresponding label sequences, avoiding the labor of labeling positions of individual characters (1). Where the training set is: $\chi = \{I_i,l_i\} $, where $I_i$ is the training image and $l_i$ is the ground truth label.

$$
\mathcal{O} = -\sum_{I_i,l_i \in \chi} \textrm{log} p(l_i|y_i)
$$

To calculate conditional probability, the input is a sequence $y = y_1, ..., y_T$ where $T$ is the number of time steps in the output sequence. Here, each $y_t$ in $R^{|L'|}$ is a probability distribution over the set $L^{'} = L \cup \{blank\}$, where $L$ contains all labels in the task, as well as a 'blank' label denoted by " ". A sequence-to-sequence mapping function $\mathcal{B}$ is defined on sequence $\pi$ in $L'^{T}$, where $T$ is the length of the sequence.  $\mathcal{B}$ maps $\pi$ onto $l$ by first removing repeated labels and the blanks. For example,  $\mathcal{B}$ maps "--jj-a-a-nn-ee--" (where '-' represents blank spaces) onto "jane". The conditional probability is defined as the sum of probabilities of all $\pi$ that are mapped by  $\mathcal{B}$ onto $l$. Under lexicon-free best-path decoding $l^*$ can be approximated as $l^\approx \mathcal{B}(\textrm{argmax}_\pi p(\pi|y))$.

\begin{equation}
p(l \mid y) = \sum_{\pi:\mathcal{B}(\pi)=l} p(\pi \mid y)
\end{equation}

The smallest possible bounding box of the area identified to contain each path of text is replaced with masking boxes with an intensity value of 0 throughout, thus masking and effectively destroying ’burned-in’ image metadata. Each unique word(s) of text is extracted and written to a respective comm-separated value file, later available to the user to re-associate with the image on an as-needed basis.

\textit{Data Preparation and Preprocessing}
All input images are uniformly scaled to a fixed height $H$, while preserving the aspect ratio. This ensures that the subsequent convolutional neural network feature extractor receives inputs of consistent height regardless of original image dimensions. No padding is applied horizontally. Thus, the width, $W$, may vary across samples, allowing the network to accommodate arbitrary-length, sequence-like objects (\textit{e.g., }words of varying length in scene text).

\textit{Convolutional Feature Extraction}
The convolutional component of our CRNN is adopted from a standard CNN architecture without its fully connected (FC) head. Convolutional and max‐pooling layers and their accompanying element‐wise nonlinearities (ReLU) are retained to extract high‐level, translation‐invariant representations from the input image. Specifically, a series of convolutional blocks, each comprising of Conv$\rightarrow$ReLU$\rightarrow$MaxPool, reduces the spatial resolution while increasing channel depth. We create feature maps of size $H \times W$, the final convolutional block outputs feature maps of size $\tfrac{H}{s} \times \tfrac{W}{s} \times C$, where $s$ is the overall downsampling factor and $C$ is the number of output channels.

To convert the 2D feature maps into a 1D sequence suitable for recurrent neural network processing, we apply a “Map-to-Sequence” layer that traverses the feature maps column by column. Each column (width = 1 pixel) across all $C$ channels is concatenated to form a feature vector $\mathbf{x}_i \in \mathbb{R}^{C}$. Thus, from a map of width $\tfrac{W}{s}$, we obtain a sequence

$$
\mathbf{x} = [\,\mathbf{x}_1, \mathbf{x}_2, \dots, \mathbf{x}_T\,],
$$

where $T = \tfrac{W}{s}$. Since convolution and pooling operations are local and translation‐invariant, the $i$-th column in the feature map corresponds to a rectangular receptive field in the original image, making each $\mathbf{x}_i$ an encoded descriptor for its associated patch.

\textit{Sequence Labeling with Bidirectional LSTM}\\
On top of the extracted feature sequence, we stack $L$ layers of bidirectional Long Short-Term Memory (BiLSTM) units. Each BiLSTM layer consists of a forward LSTM (left-to-right) and a backward LSTM (right-to-left), so each token’s representation can depend on words before and after it. These outputs at each time step, $i$, are concatenated to yield a context‐enriched hidden representation $\mathbf{h}_i$. This architecture offers: 1) Contextual modeling crucial for disambiguating visually similar symbols by leveraging surrounding context, 2) trainable feedback by allowing back-propagation of error through both the recurrent and convolutional components, enabling end-to-end optimization, and 3) length invariance by naturally handling sequences of arbitrary length $T$. Formally, at layer $l$ and time $i$:

$$
\overrightarrow{\mathbf{h}}^{(l)}_i = \mathrm{LSTM}^{(l)}_{\mathrm{fwd}}(\mathbf{h}^{(l-1)}_i, \overrightarrow{\mathbf{h}}^{(l)}_{i-1})$$

$$
\overleftarrow{\mathbf{h}}^{(l)}_i = \mathrm{LSTM}^{(l)}_{\mathrm{bwd}}(\mathbf{h}^{(l-1)}_i, \overleftarrow{\mathbf{h}}^{(l)}_{i+1}),
$$

$$
\mathbf{h}^{(l)}_i = [\,\overrightarrow{\mathbf{h}}^{(l)}_i; \overleftarrow{\mathbf{h}}^{(l)}_i\,],
$$

with $\mathbf{h}^{(0)}_i = \mathbf{x}_i$.

\textit{Transcription}\\
The recurrent layers produce a per‐time‐step label distribution $\mathbf{y}_i$ over the alphabet $\mathcal{L}\cup\{\texttt{blank}\}$. To transform this representation into a final label sequence, we employ a Connectionist Temporal Classification (CTC) loss and decoding mechanism. The CTC loss seeks to minimize the negative log‐likelihood of the ground-truth label sequence under the CTC‐defined conditional probability, which efficiently sums over all alignments via the forward–backward algorithm. During inference, either a lexicon‐free best‐path decoding (removing repeated labels and blanks) is performed to yield the recognized sequence, as we wanted to detect all the text present in the pixel data and subsequently filter with deterministic methods decreasing latency.

When training end-to-end, pixel values are scaled to $[0,1]$ and standardized per channel. We used the \texttt{Adam} optimizer with an initial learning rate of $1\times10^{-4}$, regularized with a dropout rate of 0.5, applied between BiLSTM layers. Models were trained for 100 epochs, with early stopping on validation CTC loss. All experiments were implemented in PyTorch and run on NVIDIA GPUs.

This CRNN design efficiently bridges spatial feature learning with sequence modeling, enabling robust, end-to-end recognition of sequence-like visual patterns without requiring fixed‐length inputs and character‐level bounding‐box annotations.

\subsection{Image Reconstruction via Inpainting}
The same set of synthetically created PHI DICOM images (n=29,876) was paired with each of their ground truth images before we overlaid synthetic PHI, in conjunction with the binary masks of text location following text overlay. These pairs were used to guide and assess image restoration. Three algorithms were tested: 1) TELEA algorithm\cite{telea2004image}, 2) latent-diffusion without imaging modality context, and 3) latent-diffusion with imaging modality context.

The TELEA algorithm\cite{telea2004image} was applied using OpenCV’s \texttt{cv2.inpaint} with the \texttt{INPAINT\_TELEA}
 option, an inpainting radius of 3 pixels, and the default search radius. For each corrupted image–mask pair with masked region $\Omega$ and boundary $\partial\Omega$, TELEA builds a distance map $D(x)$ using a fast marching method that approximates the Eikonal equation $|\nabla D(x)| = 1$ with boundary condition $D(x)=0$ for $x\in \partial\Omega$. Pixels on $\partial\Omega$ are inpainted first, and the algorithm proceeds inward by always selecting the unknown pixel $p\in\Omega$ with minimal $D(p)$. Its intensity is computed as a weighted average of neighbouring pixels with true or inpainted values $q$ within a radius (R=3):

$$I(p) = \frac{\displaystyle\sum_{q\in \mathcal{N}*R(p)} w(p,q), I(q)}
{\displaystyle\sum*{q\in \mathcal{N}_R(p)} w(p,q)},$$

where the weights put more empahsis on nearby pixels aligned with the local normal $\mathbf{n}(p)$ to the mask boundary, \textit{e.g.,}

$$w(p,q) \propto \frac{\max\bigl(0,(p-q)\cdot \mathbf{n}(p)\bigr)}{\lVert p-q\rVert^2}.$$

This procedure propagates structure and intensity information smoothly from $\partial\Omega$ into the masked region until all pixels in $\Omega$ are inpainted.

The two other ML-based inpainting variants implemented and testes in this workare based on a masked conditional diffusion model, namely the the sd2-community/stable-diffusion-2-inpainting model \cite{sd2_community_stable_diffusion_2_inpainting}. This model is a Stable Diffusion v2 latent diffusion variant fine-tuned for masked inpainting on 512×512 images. The model is initialized from the 512-base Stable Diffusion v2 checkpoint and further trained for $~$200k steps using LaMa-style large, irregular masks, with the masked image latent and binary mask provided as additional conditioning channels to the U-Net denoiser \cite{rombach2022high}. For our use case, we did not further finetune the model. Architecturally, the model follows the latent diffusion framework proposed by Rombach et al. \cite{rombach2022high}, where a variational autoencoder (VAE) encodes an input image ($x_0$ $\in\mathbb{R}^{H\times W\times3}$) into a lower-dimensional latent space ($z_0 \in \mathbb{R}^{H/8\times W/8\times 4}$), and a U-Net operates in this latent space while a frozen OpenCLIP text encoder provides token-level embeddings for cross-attention \cite{salimans2022progressive}. The forward (noising) process is a Markov chain
$$
q(z_t \mid z_{t-1}) = \mathcal{N}\bigl(\sqrt{1-\beta_t}, z_{t-1},, \beta_t I\bigr),
$$
which yields
$$q(z_t \mid z_0) = \mathcal{N}\bigl(\sqrt{\bar\alpha_t}, z_0,,(1-\bar\alpha_t)I\bigr),$$

with $\bar\alpha_t = \prod_{s=1}^t (1-\beta_s)$. A U-Net $\epsilon_\theta$ is trained to predict the noise ($\epsilon$) (or velocity $v$) from the noised latent $z_t$, time step $t$, and conditioning $c$ by minimizing:

$$
\mathcal{L}(\theta) = \mathbb{E}*{z_0, c, \epsilon, t}\Bigl[\bigl|\epsilon - \epsilon*\theta(z_t, t, c)\bigr|_2^2\Bigr].
$$
At inference, we integrate the reverse process using a classifier-free DDIM sampler \cite{ho2022classifier}.

All 2D medical images were resampled to 512$\times$512 and converted to a three-channel representation to match the Stable Diffusion v2 inpainting interface. Diffusion-based inpainting was performed using the Hugging Face diffusers implementation of Stable Diffusion 2 inpainting (pipeline class StableDiffusionInpaintPipeline \cite{huggingface_diffusers_stable_diffusion_pipelines}) with the sd2-community/stable-diffusion-2-inpainting weights \cite{creativeML_openrail_m_2022}. The model operates in the latent diffusion framework: a VAE encodes the input into a latent representation, a U-Net denoiser iteratively predicts noise in latent space, and text embeddings are provided via a frozen CLIP text encoder for cross-attention conditioning. For our use case, we did not further fine-tune the inpainting model.

To control the generated content within the masked region, we used the same conditioning inputs across experiments (masked image, binary mask, and fixed prompt configuration (for the 'context' variant, the text for the modality being reconstructed was used). After denoising, we enforced an identity constraint outside the inpainting mask by directly copying unmasked pixels from the original image into the final output. This compositing step guarantees that image content outside the redaction region remains unchanged, ensuring that any visual differences between input and output are confined to the masked area. All experiments complied with the CreativeML Open RAIL++-M license associated with Stable Diffusion 2 \cite{creativeML_openrail_m_2022} and used the Hugging Face diffusers library \cite{huggingface_diffusers_stable_diffusion_pipelines}.

\section{Downstream Classification Task}
Evaluating our three inpainting techniques through a downstream classification task provides an application‐driven, quantitative measure of how well each method preserves diagnostically important structures. By training a Vision Transformer–based modality classifier on the original dataset (prior to adding synthetic PHI) and then re‐evaluating it on inpainted images, we can directly observe the impact of each inpainting algorithm on the features that the model has learned to associate with different imaging modalities (CT, ultrasound, X-ray, MRI). A drop in accuracy, precision, recall, or F1 score would suggest that the inpainting method has altered or removed modality‐specific cues such as texture gradients or characteristic signal distributions, while stable or improved metrics suggest faithful reconstruction.

We compute macro‐averaged precision, recall, and F1‐score for detailed evaluation, further mitigating class‐imbalance effects, so that performance degradation in any single modality is equally weighted and visible. Logging these metrics for different inpainting approaches allows longitudinal comparison, revealing not only endpoint performance but also convergence behavior and stability during fine‐tuning. 

We employ transfer learning using a pretrained with ImageNet-21k ViT-B/16 backbone and replace its pre‐trained classification head with a new linear layer of size $4$, representing the modality classification task. Given a minibatch of $N$ images $\{x_i\}_{i=1}^N$ with one‐hot labels $\{y_i\}_{i=1}^N$, the model produces logits $z_i\in\mathbb{R}^4$, which are converted to probabilities via the softmax function, where \( k \in \{1, 2, 3, 4\} \) denotes the imaging modality; class labels given by \( 1 \rightarrow \text{magnetic resonance imaging} \), \( 2 \rightarrow \text{computed tomography} \), \( 3 \rightarrow \text{x-ray} \), and \( 4 \rightarrow \text{ultrasound} \).

$$
p_{i,k} = \frac{\exp(z_{i,k})}{\sum_{j=1}^4 \exp(z_{i,j})} \quad \text{for } k=1,\dots,4.
$$

We optimize the cross‐entropy loss

$$
\mathcal{L}_{\text{CE}} = -\frac{1}{N}\sum_{i=1}^N \sum_{k=1}^{4} y_{i,k}\,\log p_{i,k},
$$

using the Adam optimizer with learning rate $\eta = 3\times10^{-4}$.  At each training iteration, gradients $\nabla_{\theta}\mathcal{L}_{\text{CE}}$ are computed via backpropagation, parameters $\theta$ are updated, and per‐epoch averages of loss and accuracy are logged.  The model is evaluated in no‐gradient mode on the validation set at the end of each epoch, and the checkpoint with the highest validation accuracy is saved. To capture class‐balanced performance, we compute macro‐averaged precision, recall, and F1 score.

$$
\text{Precision}_{\text{macro}} = \frac{1}{4}\sum_{k=1}^4 \text{Precision}_k
$$
$$
\text{Recall}_{\text{macro}} = \frac{1}{4}\sum_{k=1}^4 \text{Recall}_k
$$
$$
F_{1,\text{macro}} = \frac{1}{4}\sum_{k=1}^4 F_{1,k}
$$

\section{Results}
We first report results on the primary objective of the proposed pipeline: the extent to which it removes PHI-bearing text from medical images. We quantify redaction effectiveness by measuring how often text within annotated PHI regions is successfully eliminated and no longer recoverable by downstream recognition or visual inspection. We subsequently report the degree to which the detected areas using various inpainting methods approximate the original image. Lastly, we report on the downstream effects these various inpainting methodologies can have in a classification-based task. 
\subsection{De-identification}
Across the held-out evaluation set (n=29,876), the automated redaction system achieved a mean F1 score of 0.892 $\pm$ 0.04, indicating good overall balance between missed and spurious redactions. Mean recall was 0.939 $\pm$ 0.06, suggesting that the vast majority of sensitive content was successfully identified and redacted, while the mean precision of 0.854 $\pm$ 0.06 indicates a moderate degree of over-redaction. Together, these metrics demonstrate that the system is tuned toward conservative de-identification, favoring protection against residual identifiers at the expense of removing some non-identifying content. Results are reported in Table \ref{tab: redact}.

\begin{table}[h]
\caption{Metrics for Binary Mask}
\centering
\begin{tabular}{ l | c }
\hline
\textbf{Metric} & \textbf{Mean $\pm$ std.}  \\
\hline
\hline
F1 score & 0.892 $\pm$ 0.04\\
Recall & 0.939 $\pm$ 0.06 \\
Precision & 0.854 $\pm$ 0.06\\
\hline
\end{tabular}
\label{tab: redact}
\end{table}

Across inpainting strategies, downstream classification performance (ascertaining the image modality e.g., chest x-ray, abdominal CT etc.) remained essentially unchanged and near-ceiling (Table~\ref{tab: downstream}). Diffusion inpainting without contextual conditioning achieved the highest scores (F1/Recall/Precision = 0.999), while diffusion with context and Telea yielded similarly high performance (all metrics = 0.998). Overall, the negligible differences ($\leq$ 0.001) indicate that the choice of inpainting method did not meaningfully impact validation F1, recall, or precision for this task.

\begin{table}[h]
\caption{Validation metrics for downstream classification task}
\centering
\begin{tabular}{ l | c| c | c }
\hline
\textbf{Inpainting Type} & \textbf{F1} & \textbf{Recall} & \textbf{Precision}\\
\hline
\hline
Diffusion (no context) & 0.999 & 0.999 & 0.999\\
Diffusion (with context) & 0.998 &  0.998 & 0.998 \\
Telea & 0.998 &  0.998 & 0.998\\
\hline
\end{tabular}
\label{tab: downstream}
\end{table}

\subsection{Inpainting Performance}
To quantitatively compare inpainting performance, we computed several error and similarity metrics restricted to the pixels within a binary evaluation mask. For each image pair, the original and inpainted images' metrics were then evaluated only over masked pixels. We calculated the sum of absolute differences, denoted $\mathrm{SAD}$, and the sum of squared differences, denoted $\mathrm{SSD}$, as global measures of total absolute and squared error within the masked region. From SSD and $N_{\text{mask}}$, we derived the mean squared error $\mathrm{MSE}$ and the root mean squared error $\mathrm{RMSE}$ as standard per-pixel error measures. Finally, we computed the structural similarity index ($\mathrm{SSIM}$), providing a perceptual similarity score between the original and inpainted images within the same masked region. These shorthand notations and associated performance metrics are recorded in Table \ref{tab:inpainting_performance}.

Across reconstruction methods, the two diffusion models (with and without context) yielded comparable pixel-wise errors, with a mean MSE of 912 $\pm$ 1586 and 880 $\pm$ 1426, respectively, and identical RMSE values of 25 $\pm$ 17 and 25 $\pm$ 16, indicating very similar fidelity at the voxel level. The Telea-based inpainting showed slightly higher error (MSE 1003 $\pm$ 1101, RMSE 28 $\pm$ 14), but substantially outperformed both diffusion approaches in structural similarity, achieving an SSIM of 0.96 $\pm$ 0.03 versus 0.83 $\pm$ 0.09 for both diffusion variants. This pattern suggests that while Telea may introduce modestly larger intensity deviations, it better preserves the global structural layout of the images, which is likely beneficial for the downstream classification task. Moreover, Diffusion models can add high-frequency detail that looks good but hurts SSIM in smaller inpainted area such as text boxes.

\begin{table}[h]
\caption{Validation metrics for downstream classification task. Recored as $\mu \ \pm \  \sigma$.}
\centering
\begin{tabular}{ l | c| c | c  }
\hline
\textbf{Metric} & \textbf{Diffusion} & \textbf{Diffusion  (NC)} & \textbf{Telea}\\
\hline
\hline
 \textbf{MSE} & 912 $\pm$ 1586 & 880 $\pm$ 1426 & 1003 $\pm$ 1101 \\
 \textbf{RMSE} & 25 $\pm$ 17 & 25 $\pm$ 16 & 28 $\pm$ 14\\
 \textbf{SSIM} & 0.83 $\pm$ 0.09 & 0.83 $\pm$ 0.09 & 0.96 $\pm$ 0.03\\
\hline
\end{tabular}
\label{tab:inpainting_performance}
\end{table}

\begin{figure}
    \centering
    \includegraphics[width=0.65\linewidth]{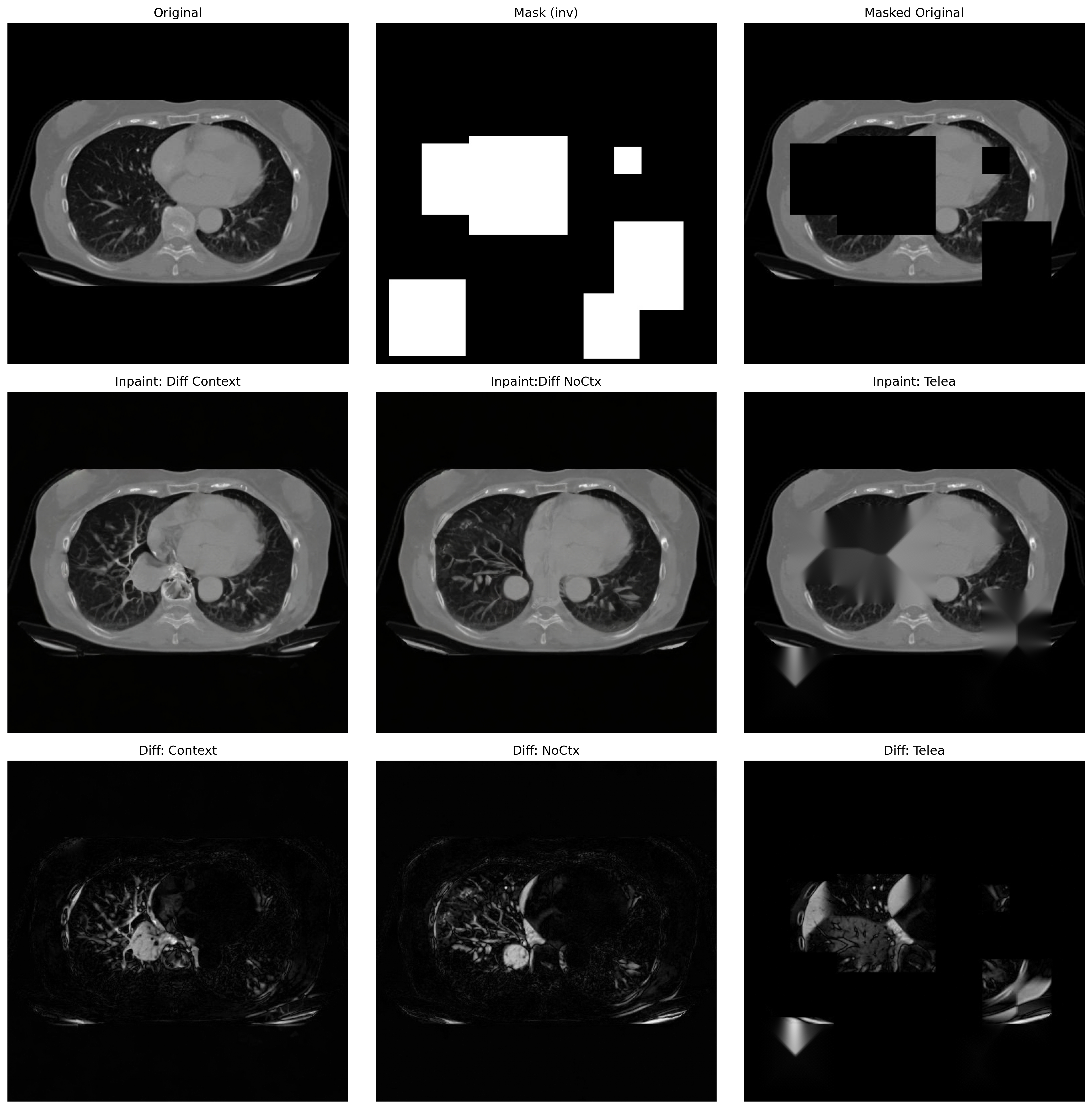}
    \caption{Masking and inpainting of 3 different techniques and their difference with the original image}
    \label{fig:enter-label}
\end{figure}

\section{Discussion}
Revisiting the aims set out in the introduction, we have developed a pipeline protocol that is robust for de-identification, sequestering relevant patient information and ROI identification. Previous work has sought to train CNNs to perform detection, and therefore removal of solely PHI-related information contained within the image, with success rates ranging from 65-89\% \cite{monteiro2017identification}\cite{huang2009privacy}\cite{lien2011open}. Some of these techniques are operating system (OS) specific or only available at a cost \cite{rodriguez2010open}. Our solution directly addresses these problems. It ensures the highly accurate removal of direct patient identifiers wihtin images while providing text file format conversion to .csv file output, and lossless image recompression. By simplifying the deep learning problem (removal/sequestering of all text versus specific text), we overcame the risk of distinguishing characters such as "8", from "B", or "0" from "o" from "O” and placed the necessary context-specific filtering on simple text filtering rather than detecting PHI directly from pixels. By extracting and subsequently masking all text, the .csv files output by the pipeline allow end users to query, include, share or destroy information as required.

Our strategy also facilitates better multi-modal integration of data information. For example, on echocardiographic images, the heart rate is often displayed as text in each view; this information is retained and made available to end users, thus is available for use during information fusion (early, joint, and late) in algorithm development \cite{kline2022multimodal}.



In the context of privacy preservation, protected health information (PHI) is first removed or masked, yielding (y) that is provably free of direct identifiers at the pixel level. Image restoration is then used to reconstruct plausible anatomy or scene structure within the masked region (1-M), while ensuring that no invertible mapping back to the original identifiers is retained. From a learning perspective, this procedure helps maintain the joint distribution over clinically relevant features, $p(x_{\text{anatomy}}, y_{\text{context}})$, in the de-identified training data. Without restoration, aggressive masking can distort this distribution. For example, by systematically removing labels, devices, or body regions in a way that encourages a model $f_\phi$ to rely on “shortcut” features $z_{\text{shortcut}}$ (such as mask boundaries, unusual blank regions, or residual PHI artifacts) instead of genuine anatomical signals $z_{\text{anat}}$. This phenomenon of shortcut learning and spurious correlations has been well documented in deep models trained on biased or systematically corrupted data.

By reconstructing masked regions so that $\hat{x}$ remains on the natural image manifold of the target domain (e.g., realistic radiographs or clinical photographs), high-quality inpainting acts to preserve the effective training distribution $p_{\text{train}}(\hat{x}$ $\approx p_{\text{real}}(x)$. This reduces the risk that the learned decision function $f_\phi$ minimizes empirical risk on artifact-ridden training data while incurring high expected risk on real-world samples,
$$\mathcal{R}*{\text{real}}(f*\phi) = \mathbb{E}*{(x,y)}\bigl[\ell(f*\phi(x),y)\bigr],$$

due to a distributional shift $p_{\text{train}}(\hat{x})$ $\neq p_{\text{real}}(x)$. In practice, restoration improves robustness and generalization by preserving structural and contextual cues—organ boundaries, device geometry, and spatial relationships that are essential for downstream tasks (classification, detection, segmentation), even when PHI has been irreversibly removed. In this way, image restoration serves as a bridge between strict privacy constraints and statistically faithful training data, allowing models to focus on clinically meaningful content rather than redaction artifacts.

Limitations to this work include that while automated data cleaning is desirable, rarely is an automated de-identification effort perfect. The risk of PHI leakage is of utmost concern due to legal and ethical ramifications. We urge the research community to test the protocol on their respective systems for ultrasound and image de-identification. Additionally, the observed image-quality trends highlight a methodological limitation: in some cases, SSIM favored the simple inpainting approach, likely because it preserves low-frequency structure and overall luminance/contrast patterns that SSIM weights heavily, even when fine textures and clinically meaningful details may be altered. As a result, higher SSIM does not necessarily imply superior clinical fidelity, and complementary metrics and expert review may be required. Another limitation is that the demonstrated downstream classification example may not reflect the complexity of many real-world clinical tasks (e.g., subtle multi-label phenotyping, rare events, domain shift across devices/sites), so performance gains may not generalize to more challenging settings. Finally, although simulated datasets are valuable for controlled evaluation, they may not fully capture the variability, artifacts, acquisition heterogeneity, and edge cases of real clinical data; thus, conclusions drawn from simulations should be interpreted cautiously and validated prospectively on diverse, multi-site datasets. Future work includes updates to the software package and incorporating feedback to make it more generalizable as adoption grows, including broader validation across institutions, devices, and clinical endpoints.

\section{Conclusion}
Oour findings demonstrate that it is possible to reconcile rigorous de-identification with preservation of anatomical fidelity by targeting only overlaid PHI while restoring the underlying signal. By coupling robust text detection with lightweight CRNN/CTC components and a latent-diffusion inpainting module, the proposed pipeline produces de-identified images that remain statistically and structurally similar to their clinical counterparts, thereby supporting downstream segmentation and classification tasks without causing distributional shifts. Future work will extend this framework to a cloud-based solution, incorporate formal re-identification risk analyses, and further explore integration with federated and privacy-preserving training paradigms to align de-identification practice with real-world clinical and regulatory constraints.


\section*{Declarations}

\section*{Funding}
Funding was not received for this work.

\section*{Conflict of Interests}
The authors declare no conflicts of interests.

\section*{Consent for publication}
All authors consent to publication of the manuscript.

\section*{Ethics approval and consent to participate:} Not applicable

\section*{Data availability:} Not applicable.
\section*{Materials availability:} Not applicable.
\section*{Clinical trial number:} Not applicable.


\FloatBarrier
\bibliographystyle{unsrt} 
\bibliography{references}  

@article{ravi2024review,
  title={A review on visual privacy preservation techniques for active and assisted living},
  author={Ravi, Siddharth and Climent-P{\'e}rez, Pau and Florez-Revuelta, Francisco},
  journal={Multimedia Tools and Applications},
  volume={83},
  number={5},
  pages={14715--14755},
  year={2024},
  publisher={Springer}
}

@misc{HHS_OCR_HIPAA_Deid_2025,
  author       = {{U.S. Department of Health and Human Services, Office for Civil Rights}},
  title        = {Guidance Regarding Methods for De-identification of Protected Health Information in Accordance with the Health Insurance Portability and Accountability Act (HIPAA) Privacy Rule},
  howpublished = {HHS.gov},
  year         = {2025},
  month        = feb,
  note         = {Content last reviewed February 3, 2025. Accessed: 2025-12-23},
  url          = {https://www.hhs.gov/hipaa/for-professionals/special-topics/de-identification/index.html}
}

@misc{melnyk2023dicomdeidvisualnlp,
  author       = {Mykola Melnyk},
  title        = {{DICOM de-identification at scale in Visual NLP (1/3)}},
  howpublished = {\emph{John Snow Labs Blog}},
  year         = {2023},
  month        = sep,
  url          = {https://www.johnsnowlabs.com/dicom-de-identification-at-scale-in-visual-nlp-1-3/},
  note         = {Accessed: 2025-12-09}
}

@article{meaney2022comparative,
  title={A comparative evaluation of transformer models for de-identification of clinical text data},
  author={Meaney, Christopher and Hakimpour, Wali and Kalia, Sumeet and Moineddin, Rahim},
  journal={arXiv preprint arXiv:2204.07056},
  year={2022}
}

@article{salimans2022progressive,
  title={Progressive distillation for fast sampling of diffusion models},
  author={Salimans, Tim and Ho, Jonathan},
  journal={arXiv preprint arXiv:2202.00512},
  year={2022}
}

@misc{sd2_community_stable_diffusion_2_inpainting,
  title        = {Stable Diffusion 2 Inpainting},
  author       = {{sd2-community}},
  howpublished = {\emph{Hugging Face model card}},
  year         = {2022},
  url          = {https://huggingface.co/sd2-community/stable-diffusion-2-inpainting},
  note         = {Accessed: 2025-12-09}
}

@misc{creativeML_openrail_m_2022,
  title        = {CreativeML OpenRAIL-M License},
  author       = {{CompVis} and {Stability AI} and {Runway}},
  year         = {2022},
  howpublished = {\emph{Stable Diffusion License Text}},
  url          = {https://huggingface.co/spaces/CompVis/stable-diffusion-license/raw/main/license.txt},
  note         = {Accessed: 2025-12-09}
}

@misc{huggingface_diffusers_stable_diffusion_pipelines,
  title        = {Stable Diffusion Pipelines},
  author       = {{Hugging Face}},
  howpublished = {\emph{Hugging Face Diffusers Documentation}},
  year         = {2025},
  url          = {https://huggingface.co/docs/diffusers/en/api/pipelines/stable_diffusion/overview},
  note         = {Accessed: 2025-12-09}
}

@article{ho2022classifier,
  title={Classifier-free diffusion guidance},
  author={Ho, Jonathan and Salimans, Tim},
  journal={arXiv preprint arXiv:2207.12598},
  year={2022}
}

@inproceedings{rombach2022high,
  title={High-resolution image synthesis with latent diffusion models},
  author={Rombach, Robin and Blattmann, Andreas and Lorenz, Dominik and Esser, Patrick and Ommer, Bj{\"o}rn},
  booktitle={Proceedings of the IEEE/CVF conference on computer vision and pattern recognition},
  pages={10684--10695},
  year={2022}
}

@article{telea2004image,
  title={An image inpainting technique based on the fast marching method},
  author={Telea, Alexandru},
  journal={Journal of graphics tools},
  volume={9},
  number={1},
  pages={23--34},
  year={2004},
  publisher={Taylor \& Francis}
}

@article{koutsoubis2024fl,
  title={Privacy preserving federated learning in medical imaging with uncertainty estimation},
  author={Koutsoubis, Nikolas and Yilmaz, Yasin and Ramachandran, Ravi P and Schabath, Matthew and Rasool, Ghulam},
  journal={arXiv preprint arXiv:2406.12815},
  year={2024}
}

@misc{johnson2024mimiccxr,
  title        = {{MIMIC-CXR} (version 2.1.0)},
  author       = {Alistair E. W. Johnson and Tom J. Pollard and Roger G. Mark
                  and Seth J. Berkowitz and Steven Horng},
  year         = {2024},
  howpublished = {\emph{PhysioNet}},
  url          = {https://physionet.org/content/mimic-cxr/2.1.0/}
}

@article{santos2025role,
  title={The Role of Deep Learning in Medical Image Inpainting: A Systematic Review},
  author={Santos, Joana Cristo and Tom{\'a}s Pereira Alexandre, Hugo and Seoane Santos, Miriam and Henriques Abreu, Pedro},
  journal={ACM Transactions on Computing for Healthcare},
  volume={6},
  number={3},
  pages={1--24},
  year={2025},
  publisher={ACM New York, NY}
}

@article{ammar2025evaluating,
  title={Evaluating the impact of different deface algorithms on deep learning segmentation software performance},
  author={Ammar, Ali and Zhu, Libing and Bryan IV, Shep and Yu, Nathan Y and Vargas, Carlos and Rong, Yi and Chen, Quan},
  journal={Frontiers in Oncology},
  volume={15},
  pages={1603593},
  year={2025}
}

@misc{wiggins2019awsdeid,
  author       = {James Wiggins},
  title        = {De-identify medical images with the help of Amazon Comprehend Medical and Amazon Rekognition},
  howpublished = {\emph{AWS Machine Learning Blog}. 
                  \href{https://aws.amazon.com/blogs/machine-learning/de-identify-medical-images-with-the-help-of-amazon-comprehend-medical-and-amazon-rekognition/}{AWS ML Blog}},
  year         = {2019},
  month        = mar,
  note         = {Accessed: 2025-12-02}

}

@article{adnan2022fldp,
  title={Federated learning and differential privacy for medical image analysis},
  author={Adnan, Mohammed and Kalra, Shivam and Cresswell, Jesse C and Taylor, Graham W and Tizhoosh, Hamid R},
  journal={Scientific reports},
  volume={12},
  number={1},
  pages={1953},
  year={2022},
  publisher={Nature Publishing Group UK London}
}

@article{packhaeuser2022reid,
  title={Deep learning-based patient re-identification is able to exploit the biometric nature of medical chest X-ray data},
  author={Packh{\"a}user, Kai and G{\"u}ndel, Sebastian and M{\"u}nster, Nicolas and Syben, Christopher and Christlein, Vincent and Maier, Andreas},
  journal={Scientific Reports},
  volume={12},
  number={1},
  pages={14851},
  year={2022},
  publisher={Nature Publishing Group UK London}
}

@inproceedings{shin2018gansynth,
  title={Medical image synthesis for data augmentation and anonymization using generative adversarial networks},
  author={Shin, Hoo-Chang and Tenenholtz, Neil A and Rogers, Jameson K and Schwarz, Christopher G and Senjem, Matthew L and Gunter, Jeffrey L and Andriole, Katherine P and Michalski, Mark},
  booktitle={International workshop on simulation and synthesis in medical imaging},
  pages={1--11},
  year={2018},
  organization={Springer}
}

@article{kondylakis2024ai4hi,
  title={Documenting the de-identification process of clinical and imaging data for AI for health imaging projects},
  author={Kondylakis, Haridimos and Catalan, Rocio and Alabart, Sara Martinez and Barelle, Caroline and Bizopoulos, Paschalis and Bobowicz, Maciej and Bona, Jonathan and Fotiadis, Dimitrios I and Garcia, Teresa and Gomez, Ignacio and others},
  journal={Insights into Imaging},
  volume={15},
  number={1},
  pages={130},
  year={2024},
  publisher={Springer}
}

@article{rempe2025deid,
  title={De-identification of medical imaging data: a comprehensive tool for ensuring patient privacy},
  author={Rempe, Moritz and Heine, Lukas and Seibold, Constantin and H{\"o}rst, Fabian and Kleesiek, Jens},
  journal={European radiology},
  pages={1--10},
  year={2025},
  publisher={Springer}
}

@inproceedings{regan2022twostage,
  title={A two-stage de-identification process for privacy-preserving medical image analysis},
  author={Shahid, Arsalan and Bazargani, Mehran H and Banahan, Paul and Mac Namee, Brian and Kechadi, Tahar and Treacy, Ceara and Regan, Gilbert and MacMahon, Peter},
  booktitle={Healthcare},
  volume={10},
  number={5},
  pages={755},
  year={2022},
  organization={MDPI}
}

@inproceedings{graves2006connectionist,
  title={Connectionist temporal classification: labelling unsegmented sequence data with recurrent neural networks},
  author={Graves, Alex and Fern{\'a}ndez, Santiago and Gomez, Faustino and Schmidhuber, J{\"u}rgen},
  booktitle={Proceedings of the 23rd international conference on Machine learning},
  pages={369--376},
  year={2006}
}

@inproceedings{wang2011end,
  title={End-to-end scene text recognition},
  author={Wang, Kai and Babenko, Boris and Belongie, Serge},
  booktitle={2011 International conference on computer vision},
  pages={1457--1464},
  year={2011},
  organization={IEEE}
}

@inproceedings{mishra2012scene,
  title={Scene text recognition using higher order language priors},
  author={Mishra, Anand and Alahari, Karteek and Jawahar, CV},
  booktitle={BMVC-British machine vision conference},
  year={2012},
  organization={BMVA}
}

@inproceedings{karatzas2013icdar,
  title={ICDAR 2013 robust reading competition},
  author={Karatzas, Dimosthenis and Shafait, Faisal and Uchida, Seiichi and Iwamura, Masakazu and i Bigorda, Lluis Gomez and Mestre, Sergi Robles and Mas, Joan and Mota, David Fernandez and Almazan, Jon Almazan and De Las Heras, Lluis Pere},
  booktitle={2013 12th international conference on document analysis and recognition},
  pages={1484--1493},
  year={2013},
  organization={IEEE}
}

@article{parker2021canadian,
  title={Canadian association of radiologists white paper on de-identification of medical imaging: part 1, general principles},
  author={Parker, William and Jaremko, Jacob L and Cicero, Mark and Azar, Marleine and El-Emam, Khaled and Gray, Bruce G and Hurrell, Casey and Lavoie-Cardinal, Flavie and Desjardins, Benoit and Lum, Andrea and others},
  journal={Canadian Association of Radiologists Journal},
  volume={72},
  number={1},
  pages={13--24},
  year={2021},
  publisher={SAGE Publications Sage CA: Los Angeles, CA}
}

@article{fezai2023deep,
  title={Deep anonymization of medical imaging},
  author={Fezai, Lobna and Urruty, Thierry and Bourdon, Pascal and Fernandez-Maloigne, Chrsitine and Alzheimer’s Disease Neuroimaging Initiative},
  journal={Multimedia Tools and Applications},
  volume={82},
  number={6},
  pages={9533--9547},
  year={2023},
  publisher={Springer}
}

@article{monteiro2017identification,
  title={A de-identification pipeline for ultrasound medical images in DICOM format},
  author={Monteiro, Eriksson and Costa, Carlos and Oliveira, Jos{\'e} Lu{\'\i}s},
  journal={Journal of medical systems},
  volume={41},
  number={5},
  pages={89},
  year={2017},
  publisher={Springer}
}

@article{huang2009privacy,
  title={Privacy preservation and information security protection for patients’ portable electronic health records},
  author={Huang, Lu-Chou and Chu, Huei-Chung and Lien, Chung-Yueh and Hsiao, Chia-Hung and Kao, Tsair},
  journal={Computers in Biology and Medicine},
  volume={39},
  number={9},
  pages={743--750},
  year={2009},
  publisher={Elsevier}
}

@misc{OAI2022,
  author       = {{National Institutes of Health (NIH)}},
  title        = {The Osteoarthritis Initiative (OAI)},
  year         = {2022},
  howpublished = {Data resource},
  url          = {https://www.nia.nih.gov/research/resource/osteoarthritis-initiative-oai},
  note         = {NIH: The Osteoarthritis Initiative (OAI)}
}

@article{shi2016end,
  title={An end-to-end trainable neural network for image-based sequence recognition and its application to scene text recognition},
  author={Shi, Baoguang and Bai, Xiang and Yao, Cong},
  journal={IEEE transactions on pattern analysis and machine intelligence},
  volume={39},
  number={11},
  pages={2298--2304},
  year={2016},
  publisher={IEEE}
}

@article{newhauser2014anonymization,
  title={Anonymization of DICOM electronic medical records for radiation therapy},
  author={Newhauser, Wayne and Jones, Timothy and Swerdloff, Stuart and Newhauser, Warren and Cilia, Mark and Carver, Robert and Halloran, Andy and Zhang, Rui},
  journal={Computers in biology and medicine},
  volume={53},
  pages={134--140},
  year={2014},
  publisher={Elsevier}
}

@article{madani2018fast,
  title={Fast and accurate view classification of echocardiograms using deep learning},
  author={Madani, Ali and Arnaout, Ramy and Mofrad, Mohammad and Arnaout, Rima},
  journal={NPJ digital medicine},
  volume={1},
  number={1},
  pages={6},
  year={2018},
  publisher={Nature Publishing Group UK London}
}

@inproceedings{kamran2019optic,
  title={Optic-net: A novel convolutional neural network for diagnosis of retinal diseases from optical tomography images},
  author={Kamran, Sharif Amit and Saha, Sourajit and Sabbir, Ali Shihab and Tavakkoli, Alireza},
  booktitle={2019 18th IEEE international conference on machine learning and applications (ICMLA)},
  pages={964--971},
  year={2019},
  organization={IEEE}
}

@article{patel2019human,
  title={Human--machine partnership with artificial intelligence for chest radiograph diagnosis},
  author={Patel, Bhavik N and Rosenberg, Louis and Willcox, Gregg and Baltaxe, David and Lyons, Mimi and Irvin, Jeremy and Rajpurkar, Pranav and Amrhein, Timothy and Gupta, Rajan and Halabi, Safwan and others},
  journal={NPJ digital medicine},
  volume={2},
  number={1},
  pages={111},
  year={2019},
  publisher={Nature Publishing Group UK London}
}

@misc{ADNI2022,
  author       = {{Alzheimer's Disease Neuroimaging Initiative (ADNI)}},
  title        = {Alzheimer's Disease Neuroimaging Initiative (ADNI)},
  year         = {2022},
  howpublished = {Data resource},
  url          = {https://adni.loni.usc.edu/},
  note         = {Alzheimer's Disease Neuroimaging Initiative (ADNI)}
}

@article{hering2022learn2reg,
  title={Learn2Reg: comprehensive multi-task medical image registration challenge, dataset and evaluation in the era of deep learning},
  author={Hering, Alessa and Hansen, Lasse and Mok, Tony CW and Chung, Albert CS and Siebert, Hanna and H{\"a}ger, Stephanie and Lange, Annkristin and Kuckertz, Sven and Heldmann, Stefan and Shao, Wei and others},
  journal={IEEE Transactions on Medical Imaging},
  volume={42},
  number={3},
  pages={697--712},
  year={2022},
  publisher={IEEE}
}

@article{biswas2019state,
  title={State-of-the-art review on deep learning in medical imaging},
  author={Biswas, Mainak and Kuppili, Venkatanareshbabu and Saba, Luca and Edla, Damodar Reddy and Suri, Harman S and Cuadrado-Godia, Elisa and Laird, John R and Marinhoe, Rui Tato and Sanches, Joao M and Nicolaides, Andrew and others},
  journal={Frontiers in Bioscience-Landmark},
  volume={24},
  number={3},
  pages={380--406},
  year={2019},
  publisher={IMR Press}
}

@article{liao2022real,
  title={Real-time scene text detection with differentiable binarization and adaptive scale fusion},
  author={Liao, Minghui and Zou, Zhisheng and Wan, Zhaoyi and Yao, Cong and Bai, Xiang},
  journal={IEEE transactions on pattern analysis and machine intelligence},
  volume={45},
  number={1},
  pages={919--931},
  year={2022},
  publisher={IEEE}
}

@inproceedings{lien2011open,
  title={Open source tools for standardized privacy protection of medical images},
  author={Lien, Chung-Yueh and Onken, Michael and Eichelberg, Marco and Kao, Tsair and Hein, Andreas},
  booktitle={Medical Imaging 2011: Advanced PACS-based Imaging Informatics and Therapeutic Applications},
  volume={7967},
  pages={177--183},
  year={2011},
  organization={SPIE}
}

@article{rodriguez2010open,
  title={An open source toolkit for medical imaging de-identification},
  author={Rodr{\'\i}guez Gonz{\'a}lez, David and Carpenter, Trevor and van Hemert, Jano I and Wardlaw, Joanna},
  journal={European radiology},
  volume={20},
  number={8},
  pages={1896--1904},
  year={2010},
  publisher={Springer}
}

@article{kline2022multimodal,
  title={Multimodal machine learning in precision health: A scoping review},
  author={Kline, Adrienne and Wang, Hanyin and Li, Yikuan and Dennis, Saya and Hutch, Meghan and Xu, Zhenxing and Wang, Fei and Cheng, Feixiong and Luo, Yuan},
  journal={NPJ digital medicine},
  volume={5},
  number={1},
  pages={171},
  year={2022},
  publisher={Nature Publishing Group UK London}
}

\newpage

\appendix

\end{document}